\documentclass[conference,a4paper]{IEEEtran}
\usepackage[T1]{fontenc}
\usepackage[utf8]{inputenc}
\usepackage{graphicx}
\usepackage{array}
\IEEEoverridecommandlockouts
\makeatletter
\def\ps@IEEEtitlepagestyle{%
  \def\@oddfoot{\mycopyrightnotice}%
  \def\@evenfoot{}%
}
\def\mycopyrightnotice{%
  {\footnotesize\parbox{\textwidth}{Accepted for presentation at the 2026 International Conference on Applied Science and Technology (iCAST-ES 2026). \copyright{} 2026 IEEE. Personal use of this material is permitted. Permission from IEEE must be obtained for all other uses, in any current or future media, including reprinting/republishing this material for advertising or promotional purposes, creating new collective works, for resale or redistribution to servers or lists, or reuse of any copyrighted component of this work in other works.}}%
  \gdef\mycopyrightnotice{}%
}
\makeatother

\begin{document}

\title{Ingest-Time Fact Compilation for Cost-Efficient and Reliable Question Answering over Revised Corpora}

\author{
\IEEEauthorblockN{Kyle Wild}
\IEEEauthorblockA{\textit{Endgame Labs, Inc.}\\
\textit{San Francisco, USA}\\
\textit{Asia AI Institute}\\
\textit{Tokyo, Japan}\\
ORCID: 0009-0001-6918-3197}
\and
\IEEEauthorblockN{Yusuke Takahashi}
\IEEEauthorblockA{\textit{Asia AI Institute,}\\
\textit{Faculty of Data Science}\\
\textit{Musashino University}\\
\textit{Tokyo, Japan}\\
ORCID: 0009-0006-8351-2280}
\and
\IEEEauthorblockN{Asako Uraki}
\IEEEauthorblockA{\textit{Asia AI Institute,}\\
\textit{Faculty of Data Science}\\
\textit{Musashino University}\\
\textit{Tokyo, Japan}\\
ORCID: 0009-0006-8412-1804}
}
\maketitle

\begin{abstract}
Most agentic question answering (``QA'') systems do an important part of their semantic work at the worst possible time: every time someone asks a question. When a corpus contains revisions, drafts, revocations, deletions, and sources with different levels of authority, the model must reconstruct the governed current state on every read---then throw that work away and repeat it on the next query. This is a bit like a database that rebuilds a materialized view every time someone reads from it. We present ingest-time fact compilation, an alternative architecture that performs this work when corpus data is ingested or changed. Raw passages are rephrased into self-contained facts; rules governing revisions, deletions, effective dates, and source trust are resolved once; and the resulting current state is stored as typed records carrying source and revision provenance. At query time, an inexpensive model reads the compiled record instead of reconstructing it from a noisy set of candidates. In a controlled synthetic experiment across five seeds, the same low-cost model produced the correct value, source, and revision in only one of 30 trials under query-time reconstruction, but in all 30 trials from the compiled substrate, at 12.89 times lower mean read cost per question. On simpler revision questions, both architectures were exact, but the compiled path used 21.6 times fewer tokens. A separate test on Federal Reserve dialogue and Wikipedia prose found that fact rephrasing roughly halved verbose dialogue while preserving high source entailment, but left the already-concise Wikipedia prose essentially unchanged. These results support a narrow but practical claim: resolving a corpus state once can make subsequent QA cheaper and more reliable for inexpensive models. We release the open source, MIT-licensed implementation and experimental artifacts so other researchers can reproduce, adapt, and extend this work.
\end{abstract}


\section{Introduction}

Organizations of every size --- companies, universities, local governments --- now maintain bodies of knowledge that change continuously: project records are corrected, policies are superseded, entries are deleted for legal reasons, and drafts coexist with approved values. Question answering (QA) over such corpora is rapidly being delegated to large language models, and whether that delegation is affordable, auditable, and fast determines where it can be deployed. Correct answers over changing institutional knowledge are an infrastructure problem: the dominant cost is not retrieval itself but the semantic labor of reconstructing governed state at query time.

The default architecture is retrieval-augmented generation (RAG) [1], [2]: retrieve raw passages, place them in the context window, and let the model derive the answer. We refer to the general pattern as query-time semantic reconstruction (QSR), because every read re-pays the same semantic labor --- deciding which of several retrieved revisions is current, which rows are drafts or sandbox artifacts, which values have been revoked or deleted, and which source should be trusted. Corpora that are revised over time make this reconstruction a governance problem, not merely a retrieval problem. Performing it inside the context window at query time is expensive in tokens and latency, degrades with context length [3], and --- as we show --- exceeds the reliable capability of inexpensive models. Prior work formalized the economics of the alternative, ingest-time semantic compilation (ISC), deriving a break-even read frequency $R^{*}$ above which compiling once and maintaining wins over reconstructing on every read, and showing that maintenance cost scales with the amount of corpus change rather than corpus size [4]. What that work did not provide is (i) a concrete fact-level compilation technique and (ii) measured evidence of the failure and cost asymmetry between the two architectures on governed, revised corpora.

In this paper, we present ingest-time fact compilation, a pipeline in which a lightweight LLM rephrases raw passages into self-contained, pronoun-free atomic facts at ingest time, governance rules are resolved once, and the results are persisted as typed, provenance-carrying substrate rows that a cheap model can read at query time. We evaluate the pipeline against QSR with a reproducible, publicly rerunnable harness over a synthetic revision corpus with explicit governance rules.

Our contributions are: (1) a fact-level compilation method that separates semantic labor (paid once, at ingest) from answering (paid on every read); (2) an open, reproducible experiment harness --- five-seed reruns of every condition plus a measured evaluation of the rephrasing step on two real source styles --- with frozen result snapshots, pinned model identifiers, and priced token accounting; (3) a failure-asymmetry finding: a low-cost model that is unreliable at governance reconstruction under QSR (mean exact rate 3.3\% across five seeds) answers perfectly from the compiled substrate in every seed; and (4) a cost-asymmetry quantification of 12.89--185.10 times mean per-question cost, with a 21.6 times reduction in tokens per question, which sets the marginal cost of governance-correct QA.

Section II positions the work. Section III defines the pipeline and its data structures. Section IV reports the experiments. Section V discusses interpretation and limitations, and Section VI concludes.

\section{Related Work}

Query-time reconstruction. RAG [1] and its descendants [2] retrieve raw evidence and reconstruct meaning in the context window. Long-context models extend how much can be reconstructed at once, but position effects degrade reliability as context grows [3], and every read still pays the full token cost. Approximate nearest-neighbor infrastructure [5] amortizes retrieval, not the per-query semantic reconstruction itself. Semantic caches [6] amortize repeated or similar queries by storing answers, but cached answers go stale under revision and carry no governance resolution.

Ingest-time structure. Databases long ago moved labor to write time: the inverted index and the materialized view are ingest-time compilations, with view maintenance studied extensively [7]. In the LLM era, GraphRAG [8] compiles a graph and community summaries at ingest. Proposition-level indexing rewrites passages into self-contained propositions to improve retrieval granularity [9], and atomic-fact decomposition is used to evaluate factual precision [10]; both share the rewriting move of our step S2 but stop at retrieval or evaluation, without governance resolution or typed provenance. Structurally, our fact row is a Slowly Changing Dimension Type 2 record [15], rules R1--R6 mirror temporal-database concerns standardized in SQL:2011 [16], and the stored source and revision fields relate to W3C PROV provenance modeling [17]; the read path S5 likewise relates to natural-language interfaces over typed stores. In the semantic-space lineage, vector-space retrieval [11], latent semantic indexing [12], and the Mathematical Model of Meaning [13] compute a semantic space once, ahead of queries --- a related precedent for paying semantic labor before read time. The ISC maintenance study [4] frames the architectural choice economically, derives the break-even read frequency $R^{*}$, and shows that a compiled substrate can be kept current under corpus change at a maintenance cost that scales with the amount of change, not corpus size.

Positioning. Two properties separate this paper from prior work: (a) compilation at the level of governance-resolved facts --- current value, revision, deletion state, and trusted source resolved once and stored with provenance --- rather than of embeddings, graphs, or summaries alone; and (b) measured failure and cost asymmetry between architectures on the same governed corpus and the same models. Prior ingest-time systems provide (a) only partially and, to our knowledge, none provide (b). In addition, similarity-only retrieval inherits the anisotropic geometry of neural embedding spaces [14], which further motivates a typed, symbol-addressable substrate.

\section{Method: Ingest-Time Fact Compilation}

\subsection{Pipeline Overview}

The pipeline has five steps, summarized in Fig. 1. S1 Segmentation: raw sources are split into passage-level units carrying their origin (path, revision identifier, environment). S2 Fact rephrasing: a lightweight LLM rewrites each unit into self-contained, pronoun-free atomic facts, so that each fact is interpretable without its surrounding document; in measured runs (Section IV-B), rephrasing compressed verbose dialogue transcripts to 0.49 times the original tokens while keeping 97.6\% of generated facts source-entailed, and left already-concise encyclopedic prose essentially unchanged (1.03 times) --- compression is source-style dependent, strongest where prose is verbose and pronoun-heavy. S3 Governance resolution: revision ordering, cross-row revocations, environment filtering, source-trust preferences, and deletion tombstones are applied once, at ingest, under the rules of Section III-C. S4 Substrate persistence: the resolved facts are stored as typed rows carrying value, source, revision, state, and effective date. S5 Query-time read: a query is answered by row lookup plus a small-model completion that cites the stored source and revision verbatim.

\subsection{Data Structures}

The unit of the substrate is a fact row: a tuple (entity, slot, value, revision, state, environment, source, effective\_on). The state field distinguishes production values, drafts, proposals, rejected rows, sandbox artifacts, revocations, and tombstones. Under QSR, the model instead receives a shuffled set of retrieved raw rows --- current and stale revisions, distractors from other entities, drafts and sandbox rows --- and must reconstruct the same tuple in the context window. The two architectures therefore answer from the same underlying information; they differ only in when the governance labor is performed and in what form it persists.

\subsection{Governance Resolution Rules}

The reference rule set, applied once at ingest (ISC) or demanded of the model at query time (QSR), is: R1 only production rows are eligible; R2 among eligible rows, the highest revision wins; R3 drafts, proposals, rejected, sandbox, and future-dated rows are ignored; R4 a revocation row that names a document identifier removes that candidate; R5 owner and regulatory sources are preferred over imported ones; R6 if the winning row is a deletion tombstone whose effective date has passed, the answer is UNKNOWN; R7 the cited source and revision must come from the chosen row itself, not from a nearby draft or sandbox row.

\subsection{Cost Model}

Following the ISC framing [4], total semantic-processing cost scales with corpus change and maintenance rather than read load once compilation is adopted: the compiled path pays a per-change ingest cost and a small maintenance term, while QSR pays a reconstruction cost on every read. The break-even read frequency $R^{*}$ falls as the per-read gap widens; the experiments below measure that gap directly on hosted models. The comparison in this paper is deliberately scoped to the marginal per-read gap; ingest-time compilation cost is characterized by the rephrasing measurements of Section IV (measured per-fact S2 ingest cost: \$0.00003 per fact on dialogue and \$0.00001 per fact on dense prose --- transparently reconstructed from the saved prompts and completions with the frozen tokenizer and the Table I prices, as the frozen runner did not retain the provider usage object), and a full empirical break-even including maintenance is follow-up work.

\begin{figure*}[!t]
\centering
\includegraphics[width=\textwidth]{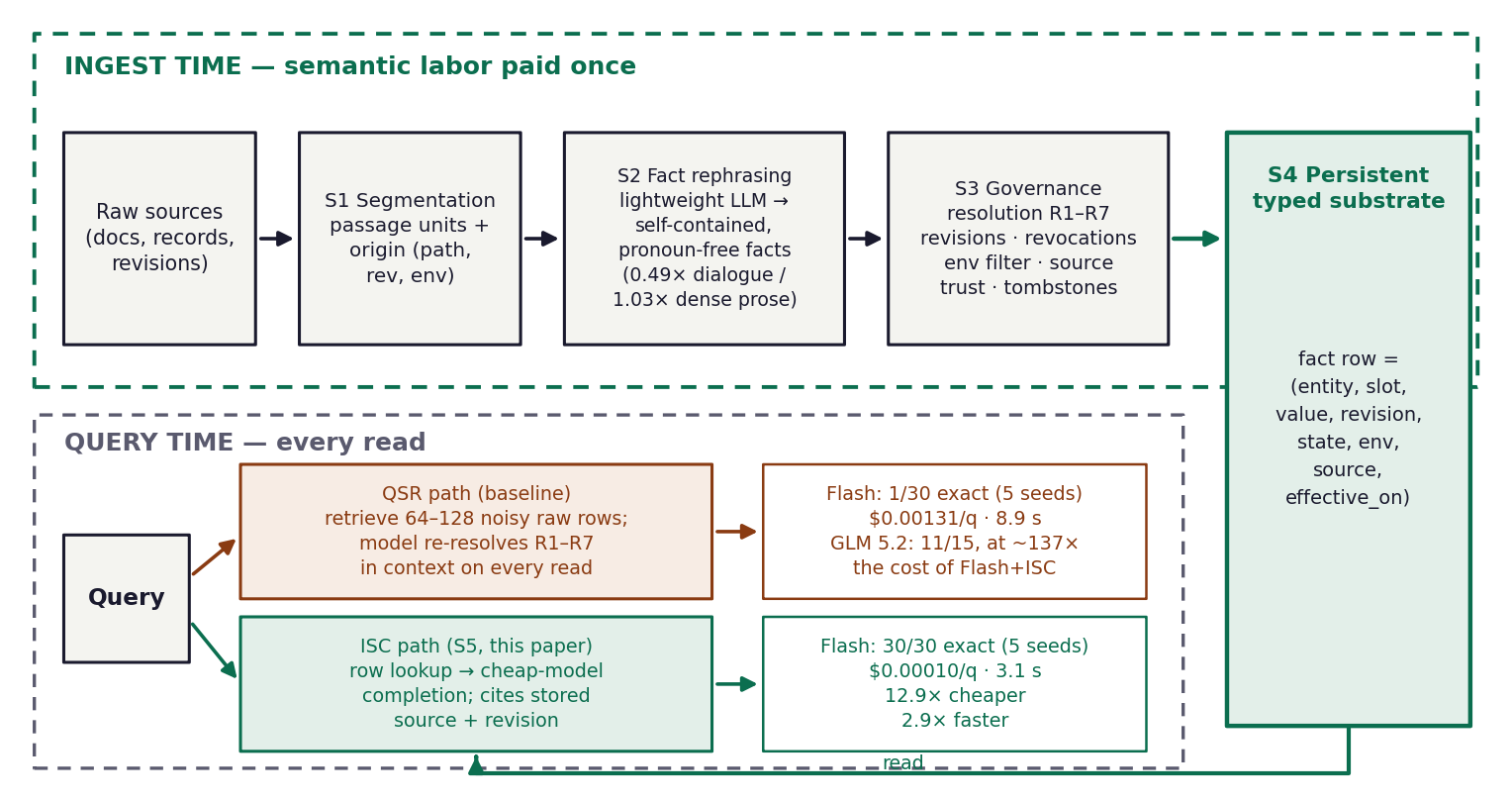}
\caption{Ingest-time fact compilation (S1--S4, paid once) and the two query-time read paths (S5 vs. QSR baseline), annotated with the measured results of Section IV.}
\label{fig1}
\end{figure*}

\section{Experiments}

\subsection{Setup}

Purpose. The evaluation isolates two questions matched to this paper's phase: does query-time governance reconstruction exceed the reliable capability of an inexpensive model while the compiled substrate does not (failure asymmetry), and what per-question cost premium does QSR pay when both architectures succeed (cost asymmetry)? External validity on real corpora is explicitly out of scope and is planned follow-up work; the results below are controlled proof-of-concept evidence. All measurements in this paper stand independently of the maintenance results reported in [4].

Harness. All runs use a Python harness against a commercial hosted API, with pinned model identifiers, JSON run configurations, and frozen result snapshots (CSV, JSONL, and manifests) dated July 8, 2026. Per-question cost is computed from logged prompt and completion tokens multiplied by the provider's published per-million-token prices as of that date (Table I); prices are recorded in the harness source for reproducibility. Corpus: a deterministic synthetic revision corpus (seed 20260708) of 80 entities, each with three slots (capital\_plan, risk\_owner, launch\_region) and three to five revisions per slot, augmented for the adversarial set with drafts, proposals, rejected rows, sandbox artifacts, future-dated rows, cross-row revocations, and deletion tombstones with effective dates. QSR conditions receive 64 (adversarial) or up to 128 (cost sweep) retrieved context rows containing the correct information plus stale revisions and distractors; ISC conditions receive the single compiled substrate row. Metric: an answer counts as exact only if the value, the cited source, and the cited revision all match. Every condition is rerun over five corpus seeds (20260709--20260713); tables report min/mean/max across seeds, and latency is the mean across seeds (latency is serving-dependent and reported for orientation only). S2 measurement: the rephrasing step is evaluated on two real passage sets --- 30 public Wikipedia passages (dense prose) and 25 valid Federal Reserve press-conference Q\&A excerpts (verbose dialogue; 5 of 30 malformed model completions excluded) --- with DeepSeek V4 Flash as the rephrasing model, tokens counted by tiktoken cl100k\_base, and three metrics: token ratio, source-entailed fact rate, and passage-level QA-fidelity pass rate. Entailment and QA-fidelity are judged automatically by a hosted LLM judge under binary criteria recorded in the harness configuration --- a fact counts as source-entailed only if fully supported by its source passage, and a passage passes QA-fidelity only if questions answered from the facts match answers from the raw passage; malformed completions are excluded from these denominators and reported (5 of 30 on the dialogue set, 16.7\%). No human agreement study accompanies these automatic judgments yet; a double-annotated validation with chance-corrected agreement is planned follow-up work. The harness, run configurations, and frozen result snapshots are available at https://github.com/aix-sc/isc (tag data-freeze-2026-07-16).

\begin{table}[!t]
\caption{Models and published prices used for cost accounting (July 8, 2026). Prices in \$ per million tokens.}
\label{tab1}
\begin{center}
\footnotesize\setlength{\tabcolsep}{3.5pt}
\begin{tabular}{|l|r|r|}
\hline
\textbf{Model} & \textbf{Input} & \textbf{Output} \\
\hline
DeepSeek V4 Flash & 0.14 & 0.28 \\
\hline
DeepSeek V4 Pro & 1.74 & 3.48 \\
\hline
GLM 5.2 & 1.40 & 4.40 \\
\hline
Kimi K2.7 Code & 0.95 & 4.00 \\
\hline
\end{tabular}
\end{center}
\end{table}

\subsection{Rephrasing Compression and Fidelity (S2)}

Table II reports the S2 measurement. On verbose dialogue, rephrasing halved the token count (0.49 times) while 97.6\% of generated facts were judged source-entailed and 92.0\% of passages passed the QA-fidelity check, with one missing major claim. On already-concise encyclopedic prose, rephrasing preserved fidelity (100.0\% entailed, 96.7\% QA fidelity, no missing claims) but did not compress (1.03 times). Compression is therefore a property of the source style, not of the pipeline alone: S2 buys its token savings where institutional text is verbose --- transcripts, meeting notes, correspondence --- and acts as a fidelity-preserving normalizer elsewhere.

\begin{table}[!t]
\caption{S2 rephrasing on two real source styles (DeepSeek V4 Flash; tokens via tiktoken cl100k\_base).}
\label{tab2}
\begin{center}
\footnotesize\setlength{\tabcolsep}{3.5pt}
\begin{tabular}{|l|r|r|r|r|}
\hline
\textbf{Source style} & \textbf{Passages} & \textbf{\shortstack[r]{Token\\ratio}} & \textbf{Entailed} & \textbf{\shortstack[r]{QA\\fidelity}} \\
\hline
Dialogue (Fed Q\&A) & 25 & 0.49$\times$ & 97.6\% & 92.0\% \\
\hline
Dense prose (Wikipedia) & 30 & 1.03$\times$ & 100.0\% & 96.7\% \\
\hline
\end{tabular}
\end{center}
\end{table}

\subsection{Adversarial Governance Reconstruction (Failure Asymmetry)}

Six questions per seed require applying rules R1--R7 to noisy context. Table III reports exact answer+source+revision accuracy, mean per-question cost, and mean latency for DeepSeek V4 Flash under both architectures across the five seeds. QSR answered at most one of six questions correctly in any seed (1/30 overall; mean exact rate 3.3\%), while the same low-cost model answered every question exactly from the compiled substrate in every seed (30/30), at 12.89 times lower mean cost and 2.9 times lower mean latency per question.

\begin{table}[!t]
\caption{Adversarial governance reconstruction with DeepSeek V4 Flash (6 questions $\times$ 5 seeds; 64 context rows for QSR). Exact as min--max across seeds; cost and latency as means.}
\label{tab3}
\begin{center}
\footnotesize\setlength{\tabcolsep}{3.5pt}
\begin{tabular}{|l|r|r|r|}
\hline
\textbf{Condition} & \textbf{\shortstack[r]{Exact\\(min--max)}} & \textbf{\shortstack[r]{Mean cost/q\\(\$)}} & \textbf{\shortstack[r]{Mean\\latency}} \\
\hline
Flash + QSR & 0/6--1/6 & 0.00131 & 8.87 s \\
\hline
Flash + ISC & 6/6 all seeds & 0.00010 & 3.06 s \\
\hline
\end{tabular}
\end{center}
\end{table}

\subsection{Heavier-Model QSR Spot Check}

To test whether a stronger model can absorb the reconstruction burden at query time, we ran a three-question sample of the same adversarial set over the five seeds (Table IV): Kimi K2.7 Code answered 0/3 to 1/3 exactly under QSR (mean 6.7\%), while GLM 5.2 partially to fully solved the sample, ranging from 2/3 to 3/3 across seeds (mean 73.3\%) --- at \$0.01389 mean per question, roughly 137 times the per-question cost of Flash reading the compiled substrate.

\begin{table}[!t]
\caption{Heavier-model QSR spot check on the adversarial sample (3 questions $\times$ 5 seeds).}
\label{tab4}
\begin{center}
\footnotesize\setlength{\tabcolsep}{3.5pt}
\begin{tabular}{|p{0.30\columnwidth}|r|r|r|}
\hline
\textbf{Condition} & \textbf{\shortstack[r]{Exact\\(min--max)}} & \textbf{\shortstack[r]{Mean cost/q\\(\$)}} & \textbf{\shortstack[r]{Mean\\latency}} \\
\hline
Kimi K2.7 Code + QSR & 0/3--1/3 & 0.01086 & 7.45 s \\
\hline
GLM 5.2 + QSR & 2/3--3/3 & 0.01389 & 7.48 s \\
\hline
\end{tabular}
\end{center}
\end{table}

\subsection{Cost Sweep on Simple Revision Tasks (Cost Asymmetry)}

On simple revision questions --- corrected facts, deletion tombstones, and ordinary current facts, without the adversarial rows --- both architectures are exact in every seed, so the comparison isolates cost (Table V). Relative to Flash + ISC, Flash + QSR at 128 context rows costs 14.21 times more and uses 21.6 times more tokens per question on average; DeepSeek V4 Pro + QSR costs 185.10 times more.

\begin{table}[!t]
\caption{Cost sweep on simple revision tasks (9 questions $\times$ 5 seeds; Pro is a 2-question sample per seed). QSR uses 128 context rows. Costs and tokens as means across seeds.}
\label{tab5}
\begin{center}
\footnotesize\setlength{\tabcolsep}{3.5pt}
\begin{tabular}{|l|l|r|r|r|}
\hline
\textbf{Condition} & \textbf{Acc.} & \textbf{\shortstack[r]{Mean cost/q\\(\$)}} & \textbf{\shortstack[r]{Mean\\tok/q}} & \textbf{\shortstack[r]{Mean\\latency}} \\
\hline
Flash + ISC & all exact & 0.00010 & 453 & 3.30 s \\
\hline
Flash + QSR & all exact & 0.00142 & 9,777 & 3.40 s \\
\hline
Pro + QSR & all exact & 0.01851 & 10,018 & 10.67 s \\
\hline
\end{tabular}
\end{center}
\end{table}

\section{Discussion}

Claims and evidence. Contribution (3) is supported by Table III: on governance-heavy reconstruction, the inexpensive model succeeds only when the governance labor has already been compiled --- in every seed. Contribution (4) is supported by Tables III--V: when both paths succeed, QSR pays a 14.21--185.10 times mean per-question cost premium and a 21.6 times token premium. Contribution (1) is qualified by Table II: the rephrasing stage compresses where the source is verbose and preserves fidelity elsewhere. Together they suggest a capability-threshold reading: query-time reconstruction demands a model above some capability level on every read, whereas compilation pays that level once at ingest and lets every subsequent read run on a cheap model. Decomposing the 29 failed QSR answers sharpens this reading: none produced the correct value with a misattributed citation --- all 29 failed to recover the correct value at all --- so reconstruction fails at value recovery, not merely at citation discipline (per-question logs in the released artifact). A heavier model (GLM 5.2) can partially substitute capability for compilation --- 2/3 to 3/3 across seeds --- but at two orders of magnitude higher marginal cost and still without seed-robust exactness.

Cost and token footprint. Tokens processed per question fall by a factor of 21.6 on average under the compiled path with identical accuracy in the cost sweep. Because inference cost, latency, and energy all scale with tokens processed, the compiled path lowers the recurring marginal cost of governance-correct QA by one to two orders of magnitude, moving routine deployment within reach of commodity models and modest serving budgets in this controlled setup.

A deterministic query-time filtering baseline. Since R1--R7 are declarative predicates over typed fields, a practitioner might filter on state and effective\_on and take the highest revision in ordinary code --- the non-materialized-view counterpart of [7] --- rather than hand 64 unfiltered rows to a model. Such a filter, however, presupposes exactly the typed, governance-relevant fields that ingest-time compilation produces: on raw revised sources, state, revision, revocation targets, and effective dates are latent in prose, and extracting them is the semantic labor S1--S4 pays once. Once rows are typed, deterministic evaluation is indeed the right reader --- the ISC read path approximates it with a near-deterministic lookup plus a small completion --- so the two designs are complements rather than competitors. A measured deterministic arm over the compiled substrate is added to the follow-up comparison plan.

Error persistence. At 97.6\% entailment, roughly one stored fact in forty is not supported by its source yet is served with provenance on every read --- a failure mode query-time reconstruction avoids by re-reading sources. Planned countermeasures are audit sampling of stored rows against sources, read-time entailment spot checks on low-confidence rows, and provenance-preserving re-verification within the maintenance cycle of [4]; until these are validated, ``reliable'' in our title should be read as governance-rule reliability rather than source-fidelity certainty.

Limitations. First, the harness is synthetic: it isolates failure modes and cost mechanics under controlled governance rules, and it supports no claim of external validity on real corpora. Second, the question sets are small (6, 3, and 9 questions per seed). Pooling runs across the five seeds --- the questions repeat per seed, so pooling tests run-level consistency rather than question-population generality --- the adversarial exact-rate difference is nevertheless decisive (Fisher's exact test, 1/30 vs 30/30, p $\approx$ $2.3\times10^{-13}$; Wilson 95\% intervals 0.6--16.7\% for QSR and 88.6--100\% for ISC); results should still be read as a controlled demonstration, not a population estimate. Third, prices are one provider's published rates on one date; ratios will vary with providers and time even though the token asymmetry is architectural. Fourth, the S2 rephrasing measurements of Section IV cover two source styles, a single rephrasing model, and a single tokenizer, and their entailment and QA-fidelity judgments are automatic, with no human agreement study yet; broader styles, domains, and human validation remain future work. Fifth, substrate maintenance under corpus change is treated by the companion maintenance work [4] and is not re-measured here.

\section{Conclusion and Future Work}

We presented ingest-time fact compilation, a pipeline that pays governance-resolving semantic labor once, at ingest, instead of upon every query. The system persists provenance-carrying fact rows, and we measured its failure and cost asymmetry against query-time semantic reconstruction on hosted models over five seeds: a 3.3\% mean exact rate versus 100\% in every seed on adversarial governance questions for the same inexpensive model, a 12.89--185.10 times mean per-question cost premium for reconstruction, and a rephrasing stage that halves tokens on verbose dialogue while preserving fidelity on concise prose. The next steps are concrete: (i) reproduce the protocol on a real revision stream --- e.g., customer support transcripts, sales call transcripts, or biographical interviews over time --- with equivalent governance rules; (ii) harden the adversarial set with multi-hop updates, conflicting currency flags, and access-control policies; (iii) extend the S2 measurement across further source styles and domains and reduce the malformed-completion rate; and (iv) integrate the incremental maintenance mechanism of [4] so that compiled substrates remain current under continuous change.

\section*{Acknowledgment}

The authors used an AI assistant (Anthropic Claude) for language editing and figure preparation. All technical content, experimental design, measurements, and claims are the authors' own, and the authors take full responsibility for the content of this paper. Part of this work used computational infrastructure provided by Endgame Labs, Inc.


\begin{thebibliography}{17}

\bibitem{ref1} P. Lewis et al., ``Retrieval-augmented generation for knowledge-intensive NLP tasks,'' in Adv. Neural Inf. Process. Syst., vol. 33, 2020, pp. 9459--9474.

\bibitem{ref2} Y. Gao et al., ``Retrieval-augmented generation for large language models: A survey,'' arXiv preprint arXiv:2312.10997, 2023.

\bibitem{ref3} N. F. Liu et al., ``Lost in the middle: How language models use long contexts,'' Trans. Assoc. Comput. Linguist., vol. 12, pp. 157--173, 2024.

\bibitem{ref4} Y. Takahashi, K. Wild, and A. Uraki, ``Cost scales with change, not corpus size: Incrementally maintaining an evolving semantic substrate,'' in Proc. Int. Electron. Symp. (IES-KCIC), Aug. 1--3, 2026, to appear. arXiv:2608.16621, doi: 10.48550/arXiv.2608.16621.

\bibitem{ref5} J. Johnson, M. Douze, and H. Jégou, ``Billion-scale similarity search with GPUs,'' IEEE Trans. Big Data, vol. 7, no. 3, pp. 535--547, 2021.

\bibitem{ref6} F. Bang, ``GPTCache: An open-source semantic cache for LLM applications,'' in Proc. NLP-OSS Workshop, 2023, pp. 212--218.

\bibitem{ref7} A. Gupta and I. S. Mumick, ``Maintenance of materialized views: Problems, techniques, and applications,'' IEEE Data Eng. Bull., vol. 18, no. 2, pp. 3--18, 1995.

\bibitem{ref8} D. Edge et al., ``From local to global: A graph RAG approach to query-focused summarization,'' arXiv preprint arXiv:2404.16130, 2024.

\bibitem{ref9} T. Chen et al., ``Dense X retrieval: What retrieval granularity should we use?'' in Proc. EMNLP, 2024, pp. 15159--15177.

\bibitem{ref10} S. Min et al., ``FActScore: Fine-grained atomic evaluation of factual precision in long form text generation,'' in Proc. EMNLP, 2023, pp. 12076--12100.

\bibitem{ref11} G. Salton and M. J. McGill, Introduction to Modern Information Retrieval. New York, NY, USA: McGraw-Hill, 1983.

\bibitem{ref12} S. Deerwester, S. T. Dumais, G. W. Furnas, T. K. Landauer, and R. Harshman, ``Indexing by latent semantic analysis,'' J. Am. Soc. Inf. Sci., vol. 41, no. 6, pp. 391--407, 1990.

\bibitem{ref13} Y. Kiyoki, T. Kitagawa, and T. Hayama, ``A metadatabase system for semantic image search by a mathematical model of meaning,'' ACM SIGMOD Record, vol. 23, no. 4, pp. 34--41, 1994.

\bibitem{ref14} K. Ethayarajh, ``How contextual are contextualized word representations? Comparing the geometry of BERT, ELMo, and GPT-2 embeddings,'' in Proc. EMNLP-IJCNLP, 2019, pp. 55--65.

\bibitem{ref15} R. Kimball and M. Ross, The Data Warehouse Toolkit: The Definitive Guide to Dimensional Modeling, 3rd ed. Indianapolis, IN, USA: Wiley, 2013.

\bibitem{ref16} K. Kulkarni and J.-E. Michels, ``Temporal features in SQL:2011,'' ACM SIGMOD Rec., vol. 41, no. 3, pp. 34--43, 2012.

\bibitem{ref17} L. Moreau and P. Missier, Eds., ``PROV-DM: The PROV data model,'' W3C Recommendation, Apr. 2013.

\end{thebibliography}
\end{document}